\documentclass[final,5p,times,twocolumn]{elsarticle}

\usepackage{mathtools,amsmath,amssymb}
\usepackage{stmaryrd}
\usepackage[hyphens]{url}
\usepackage{paralist}
\usepackage{graphicx}
\usepackage{rotating}
\usepackage{boldline}
\usepackage{framed,enumerate, soul, xcolor}
\usepackage[T1]{fontenc}
\usepackage{multirow}
\usepackage{parskip,textcomp}
\usepackage{subcaption}
\begin{document}
%%%%%%%%%%%%%%%%%%%%%%%%%%%%%%%%%%%%%%%%%%%%%%%%%%%%%%%%%%%%%%%%
%%%%%%%%%%%%%%%%%%%%%%%%%%%%%%%%%%%%%%%%%%%%%%%%%%%%%%%%%%%%%%%%
\begin{frontmatter}

%% Title
\title{Ontology Design Facilitating Wikibase Integration -- and a Worked Example for Historical Data}

%% Authors
\author[ks]{Cogan Shimizu}
\ead{cogan.shimizu@coganshimizu.com}

\author[ks]{Andrew Eells}
\ead{andreweells@ksu.edu}

\author[msu]{Seila Gonzalez}
\ead{seilagonzaleze@gmail.com}

\author[ks]{Lu Zhou}
\ead{luzhou@ksu.edu}

\author[ks]{Pascal Hitzler}
\ead{hitzler@ksu.edu}

\author[msu]{Alicia Sheill}
\ead{asheill@msu.edu}

\author[msu]{Catherine Foley}
\ead{foleyc@msu.edu}

\author[msu]{Dean Rehberger}
\ead{rehberge@msu.edu}

%% Affils
\address[ks]{Department of Computer Science, Kansas State University, USA}
\address[msu]{MATRIX, Michigan State University, USA}

%% Abstract
\begin{abstract}
Wikibase -- which is the software underlying Wikidata -- is a powerful platform for knowledge graph creation and management. However, it has been developed with a crowd-sourced knowledge graph creation scenario in mind, which in particular means that it has not been designed for use case scenarios in which a tightly controlled high-quality schema, in the form of an ontology, is to be imposed, and indeed, independently developed ontologies do not necessarily map seamlessly to the Wikibase approach. In this paper, we provide the key ingredients needed in order to combine traditional ontology modeling with use of the Wikibase platform, namely a set of \emph{axiom} patterns that bridge the paradigm gap, together with usage instructions and a worked example for historical data.
\end{abstract}

%% Keywords

\end{frontmatter}

%%%%%%%%%%%%%%%%%%%%%%%%%%%%%%%%%%%%%%%%%%%%%%%%%%%%%%%%%%%%%%%%
\section{Introduction}
\label{sec:intro}
% Primary: Pascal
%%%%%%
When developing a knowledge graph, there are many aspects to consider during its deployment. These range from usability of its interfaces (both human and programmatic), the (re)usability of the data that it contains, its accessibility (both in terms of uptime and its interfaces), transparency (relating to provenance and trustworthiness), and its persistence (preventing link rot). These characteristics are neatly summarized in the FAIR manifesto \cite{fair}. One way of accomplishing (re)usability of the data is through the principled use (i.e., using a structured development methodology) of a schema that describes and documents the relations between concepts in the knowledge graph. With respect to accessibility and persistence, one can consider exposing a SPARQL endpoint and allowing interested parties to query against it. While this is a very flexible approach, it makes it difficult to explore the data. On the other hand, one could consider exposing data through a framework such as Wikibase. In this paper, we explore how the Modular Ontology Modeling methodology (MOMo \cite{momo-swj}) can be applied in such a way that eventual deployment of the graph data to the Wikibase model is seamless. 

% Talk about MOMo & Patterns
Modular Ontology Modeling (MOMo) specifies the development of ontology modules for sets of tightly bound key notions that will be included in a given ontology \cite{momo-swj}.\footnote{We focus on the MOMo paradigm as it is closely aligned with our use case, but any pattern-based methodology, such as eXtreme Design \cite{extreme1} would work similarly.} When developing a module, it is generally suggested to identify applicable \emph{ontology design patterns} (ODPs) \cite{odp1} and adapt them to the use-case at hand through template-based instantiation \cite{template}. During this process it is good practice to consult existing collections of ODPs, such as those on the ODP Community Wiki\footnote{See \url{https://ontologydesignpatterns.org/}.} or in the MODL library \cite{modl}.

% What is Wikibase and why is it important?
As one of the largest publicly editable and accessible knowledge bases, Wikidata is an immense, crowd-sourced knowledge base with persistent data that is available for public use and consumption. Wikidata contains millions of pieces of knowledge from many different domains in the world, and is growing constantly. In addition, it serves as the structured data hub for all of Wikimedia's projects (e.g., Wikipedia, Wikivoyage, Wiktionary, and Wikisource). As such, when modeling an ontology, it makes sense to consider ease of integration with resources like Wikidata, among other Linked Data Platforms.\footnote{See \url{https://www.wikidata.org/}.} \emph{Wikibase} is the software underlying Wikidata, which can be used separately from Wikidata for knowledge graph creation and management.

% What is our motivating use case?
The Enslaved Ontology \cite{enslaved} was modeled using the nascent MOMo methodology and is used as the schema for the knowledge graph that underlies the publicly available knowledge base on the Enslaved Hub.\footnote{\url{https://enslaved.org/}} The Enslaved Hub is an innovative and compelling centralized location for engaging with historical slave trade data from a variety of sources and is supported by an underlying installation of the Wikibase platform.
During development, we had assumed that it would be relatively easy to adapt the modular Enslaved ontology to the Wikibase model. Unfortunately, between different semantics for validating data to be put into the knowledge base and unclear mapping between different notions of provenance, it was not as straightforward as we had expected, resulting in a realization of the knowledge graph in Wikibase that is conceptually close but still markedly different from the designed Enslaved Ontology \cite{DBLP:conf/cikm/ZhouSHSEFTR20}. However, during the course of that work we realized how the gap between ODP-based ontology modeling and Wikibase software use can be closed, centrally by basing the ontology design on new ODPs that are developed to allow for seamless use with Wikibase. The result of our subsequent work is what we present in this paper.

So in this paper, we will present a library of ontology design patterns that have been specifically engineered to explicitly represent how Wikibase models data ``under-the-hood,'' thus ensuring that the ontology is optimally structured for interoperability with Wikibase. This library can be used by any organization to model their own internal and proprietary knowledge graphs and apply their alignments to Wikibase as an important tool to augment or induce new information into their own knowledge graph. In particular, to the best of our knowledge this paper provides the first ODP library that provides a bridge between traditional ontology engineering and use of the Wikibase platform. 

%%%%%%%%%%%%%%%%%%%%%%%%%%%%%%%%%%%%%%%%%%%%%%%%%%%%%%%%%%%%%%%%
% Directory Information
The rest of this paper is organized as follows. In Section \ref{sec:motivation} we describe relevant aspects of Wikibase and how they give rise to mismatches with traditional ontology modeling. In Section \ref{sec:patterns} we describe our ODP library and how it addresses these mismatches. In Section \ref{sec:example} we provide a case study: a reconstructed Enslaved Ontology. Section \ref{sec:related} discusses related work before we conclude in Section \ref{sec:conc}. 

This paper is a very substantial extension of work previously presented at a workshop \cite{wikibase-wop}. Our pattern library is publicly available from our online portal.\footnote{\url{https://gitlab.cs.ksu.edu/daselab/wikibase-ontology-design-library}}

%%%%%%%%%%%%%%%%%%%%%%%%%%%%%%%%%%%%%%%%%%%%%%%%%%%%%%%%%%%%%%%%
\section{Background \& Motivation}
\label{sec:motivation}
%%%%%
% Primary: Pascal
%%%%%
\begin{figure}[tb]
    \centering
    \includegraphics[width=\columnwidth]{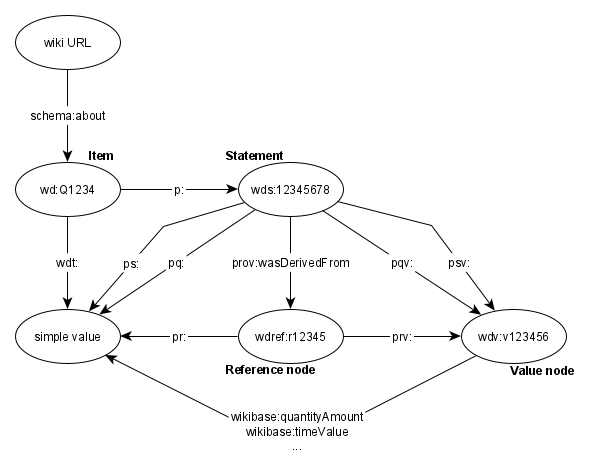}
    \caption{Wikibase RDF export schematic.}
    \label{fig:wb-rdf}
\end{figure}
%%%%%

%%%%%%%%%%%%%%%%%%%%%%%%%%%%%%%%%%%%%%%%%%%%%%%%%%%%%%%%%%%%%%%%
\subsection{The Wikibase Model}
%%%%%
%% Describe the Wikibase Model
Briefly, the Wikibase RDF export model (Figure~\ref{fig:wb-rdf})\footnote{This is a redrawing of the figure at \url{https://www.mediawiki.org/wiki/Wikibase/Indexing/RDF_Dump_Format}, where more information can be found. Further details are in \cite{DBLP:conf/semweb/ErxlebenGKMV14}.} uses the notion of reification to attach \emph{qualifiers} and \emph{references} (for provenance) to assertional statements. Reification is the practice of turning a property into an instance, so that additional assertions may be attached to the property. A common use is to attach a temporal scope to a role; for example, an employee is employed at a company from 2009-2014. In Wikibase terminology, this temporal scope would be considered a qualifier. The supporting documentation, perhaps digital tax records, could be considered a reference for this information.

%%%%%
\begin{figure}[tb]
    \centering
    \includegraphics[width=\columnwidth]{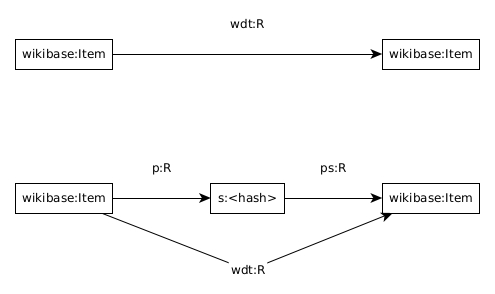}
    \caption{A closer look at the reification of \textsf{wdt:R} \textcolor{blue}{polish figure}.}
    \label{fig:wb-ex}
\end{figure}
%%%%%

Wikibase goes on to simplify how this reification occurs. The assertion is hashed, which is then turned into an instance of \textsf{wikibase:Statement}. Then, the property name is reused in two different namespaces, to complete the reification. To continue our example,%
% A note on internal identifiers Q... and P... and displays via labels
\footnote{For readability's sake, we do not use in the example the internal Wikibase identifiers \textsf{QXXXX} and \textsf{PXXXX}. The human readable representations found in the interface are from \textsf{rdfs:label} values.} %
it would be serialized in RDF as
\begin{align*}
&\textsf{ex:employee0} && \textsf{p:hasJob}   && \textsf{s:12345}   && \textsf{ .} \\
&\textsf{s:12345}      && \textsf{ps:hasJob}  && \textsf{ex:job0}   && \textsf{ .} \\
&\textsf{ex:employee0} && \textsf{wdt:hasJob} && \textsf{ex:job0}   && \textsf{ .} 
\end{align*}
where \textsf{s:12345} is the hash node created by the system. This hashing is diagrammatically shown in Figure \ref{fig:wb-ex}. In Figure \ref{fig:wb-rdf} the corresponding nodes are Item, Statement and the simple value; note the use of prefixes. It will become clear from our discussion below in Section \ref{sec:patterns} how exactly this RDF export is produced.

%%%%%%%%%%%%%%%%%%%%%%%%%%%%%%%%%%%%%%%%%%%%%%%%%%%%%%%%%%%%%%%%
\subsection{Paradigmatic Conflicts}
\label{ssec:mismatches}
%%%%%
%% Mismatches between ontology and wikibase.
% RHS Role Chains are not allowed.

The Wikibase approach to creating an RDF graph puts limitations on the graph structures that can be created through Wikibase. Some of these restrictions are mild in terms of graph modeling, however others forbid graph structures that are sometimes desirable. These restrictions primarily come from the fact that the Wikibase approach restrains what can be stated about the reification nodes (i.e., the hashes). We give some examples from the Enslaved ontology.

%%%%%
\begin{figure}[tb]
    \centering
    \includegraphics[width=.75\columnwidth]{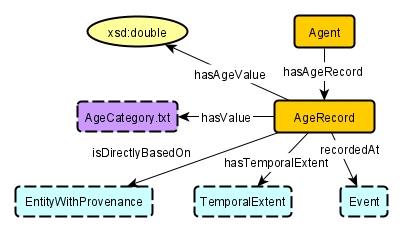}
    \caption{Enslaved Ontology: \textsf{AgeRecord} representation.}
    \label{fig:e-arc}
\end{figure}
%%%%%

Figure \ref{fig:e-arc} depicts a part of a schema diagram for the AgeRecord module in the Enslaved ontology. The color and shape coding can be mostly ignored for this discussion: boxes represent classes (concepts), ovals represent datatype values and arrows represent binary relationships (properties). Conceptually, this is to represent information about the age of an \textsf{Agent} (person) at a specific point in time. The provenance information (\textsf{isDirectlyBasedOn} relation) is for representing the origin of the data in the record. Clearly, the lower row (light blue) are references or qualifiers (using the Wikibase terminology). 

The upper left of the diagram is what presents the difficulty. The schema is based on the fact that some sources in this application context report age as a number, while others present age categories, like \emph{child} or \emph{infant}, in this application case, as a controlled vocabulary. From a (faithful) data integration perspective, the desire in this case was to make it possible to record either or both types of information. 

Now, if we take the light blue boxes as qualifiers and references, then the \textsf{AgeRecord} node would be the hashed node in the Wikibase system. However we would not be able to have both the number value and the controlled vocabulary attached to the hash node \emph{on par}. The Wikibase approach would force us to pick one of them (e.g., the controlled vocabulary) as the primary relation for the age, while the other (the number in this case) would be another qualifier. While this is \emph{possible}, it is arguably not a faithful representation of the conceptual model which does not (and should not, in this case) prefer one way of recording age over the other.

%%%%%
\begin{figure}[tb]
    \centering
    \includegraphics[width=.8\columnwidth]{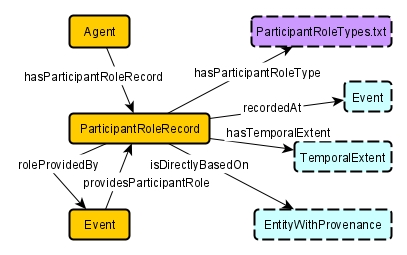}
    \caption{Enslaved Ontology: \textsf{ParticipantRoleRecord} representation.}
    \label{fig:e-prrc}
\end{figure}
%%%%%

Figure \ref{fig:e-prrc} shows another Enslaved schema diagram, in this case for historical records on participation in an event. The participant role type (intended use with a controlled vocabulary) indicates the role of the participant, e.g., the role of captain in a slave transport voyage. As before, the light blue boxes are most naturally taken as qualifiers or references. And again, as before, we are left with a decision as to the main relationship, and different arguments can be made. For example, from an agent-centric perspective, we would argue that the relation between \textsf{Agent} and \textsf{Event} is primary, in which case the type of participation would be relegated to a qualifier. However, we may also argue that the type of participation (e.g., whether as enslaved person or buyer in a sale) is of most interest in some situations, in which case the relation between \textsf{Agent} and the participation type is primary, and the event modeled as a qualifier. While either is possible, in principle, we note that the Wikibase model forces to pick a primary relationship, while the original intention for the ontology was to remain impartial in this respect. 

Another complication arises from the decision -- on the side of the ontology -- to include an inverse role between \textsf{Event} and \textsf{ParticipantRoleRecord}, in this particular case. Since \textsf{ParticipantRoleRecord} will end up being the hash node, this cannot be done in Wikibase. We are forced to use a particular direction, in this case the \textsf{roleProvidedBy} relation, as we have limited control over relation directions involving hash nodes.

%%%%%
\begin{figure}[tb]
    \centering
    \includegraphics[width=\columnwidth]{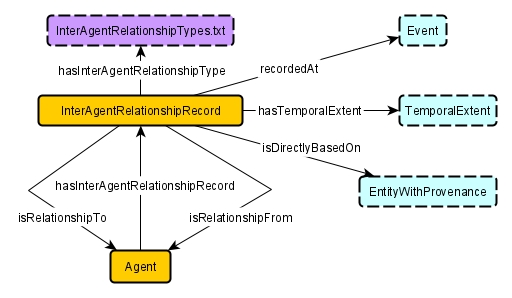}
    \caption{Enslaved Ontology: \textsf{InterAgentRelationshipRecord} representation.}
    \label{fig:e-iarrc}
\end{figure}
%%%%%

The situation depicted in Figure \ref{fig:e-iarrc} is very similar to the one just discussed; in this case it is about recording relationships \emph{between} agents (persons). For Wikibase, we are forced to make a determination whether the agent-agent relation is primary, or whether the type of relationship (as viewed from one of the agents) is primary. In addition, there are several options how to resolve the choice in the ontology model to have three relations between agents and the (reified) record; in particular the symmetry in the ontology in terms of \textsf{isRelationshipFrom} and \textsf{isRelationshipTo} cannot be maintained.

%%%%%
\begin{figure}[tb]
    \centering
    \includegraphics[width=\columnwidth]{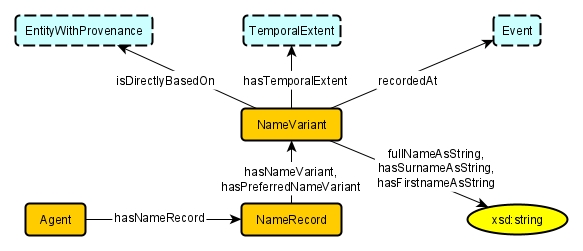}
    \caption{Enslaved Ontology: \textsf{NameRecord} representation.}
    \label{fig:e-nrc}
\end{figure}
%%%%%

Figure \ref{fig:e-nrc} is an example of a different type, the intended use is for recording names and name variants of persons. The most natural candidates for qualifiers and references are again the blue boxes, while the most natural candidate for the primary relation is probably between \textsf{Agent} and the string, but we will have to pick on of the three relations between \textsf{NameVariant} and the string as the primary relation and relegate the other two to qualifiers. In this case, however, we will not be able to have two nodes between \textsf{Agent} and the string, as Wikibase produces only one hash node. The natural grouping that the ontology provides is thus lost, unless one takes the relation between \textsf{NameRecord} and the string as the primary relation, which does not come naturally from a modeling perspective. Another attempt to force this into the Wikibase RDF model may be to make name variants qualifiers to the hash node; however since we cannot provide qualifiers to qualifier relations (or to reference relations), this would prohibit us from having, for example, provenance information for name variants.

Regarding axiomatization of an underlying ontology, the Wikibase approach also imposes some restrictions, and we will discuss these issues in more detail later in the paper. 

% No Semantics over property graphs in OWL
Another method of describing the Wikibase model succinctly would be to use a property graph \cite{prop-graphs}, such as through the use of RDF* and SPARQL*. However, there is currently not a method for axiomatically describing these sorts of graphs.\footnote{In our search we did find a corresponding OWL*, \url{https://github.com/cmungall/owlstar} however, it is still in a draft prototype phase.}

It may need emphasizing that we list the above limitations of the Wikibase approach \emph{not} in order to find fault in it. Rather, the RDF export realization appears to be a very clever use of namespaces for the serialization of reification. Most importantly, though, the Wikibase approach was put in place for a very clearly defined use case, namely to support the crowd-sourced development of Wikidata. As such, it restricts the freedom of the user in choosing graph structure in both explicit and implicit ways, by prompting the user to think (and structure the data) in terms of primary relations with attached qualifiers and references. This appears to be a very natural and, in many cases, very adequate approach. However, these advantages also come with the disadvantage of a certain loss of flexibility, in particular pertaining to the representation of more strongly structured and more fine-grained data.

Now, in application cases where stronger demands are made on graph structure -- and as we have seen from the examples above -- we cannot readily or easily transfer an ontology-based schema into the Wikibase format. But rather than having to choose between an ontology modeling approach or use of Wikibase, in this paper we will show how we can make them work together more seamlessly, by providing ontology design patterns that capture -- and thus cater for -- the restrictions imposed by Wikibase.

%%%%%%%%%%%%%%%%%%%%%%%%%%%%%%%%%%%%%%%%%%%%%%%%%%%%%%%%%%%%%%%%
\section{Generic Wikibase Patterns}
\label{sec:patterns}
%%%%%
% Primary: Cogan
% Essentially, the patterns from the WOP paper, explained. 
%%%%%
% Core
There are two core deliverables in this manuscript. First is a library of \emph{conceptual} design patterns that visually represent a paradigm for developing a schema that will immediately align to the Wikidata model. The patterns are designed to be intuitive by condensing, or folding, many of the Wikibase particulars away from the interested developer, which at the same time will prompt the developer to think about the schema in terms of primary relations with attached qualifiers and references, as discussed above. The second deliverable is the set of \emph{expanded} patterns that conform to the structure of the Wikibase RDF export, and which are capable of being axiomatically described.\footnote{This has a few caveats which are discussed individually in the following subsections.}

These patterns are presented in detail, paired together, in Figures~\ref{fig:pattern-generic}-\ref{fig:state-s} and discussed in Section~\ref{ssec:axioms}.
% Graphical Syntax & Condensing and Expansion
The diagrams and their syntax are summarized in the next sections (Sections \ref{ssec:sds} and \ref{ssec:syntax}), together with a discussion what it means to expand the conceptual diagrams.
% Shapes
We also provide the artifacts associated with this manuscript: a way to validate the resulting serializations of the expanded patterns, specified in ShEx (the Shape Expression Language\footnote{https://shex.io/}) and WOPL, the Wikibase Ontology Design Pattern Library, which follows the MODL architecture \cite{modl} in Section~\ref{ssec:resources}.

%%%%%%%%%%%%%%%%%%%%%%%%%%%%%%%%%%%%%%%%%%%%%%%%%%%%%%%%%%%%%%%%%
\subsection{Schema Diagrams}
\label{ssec:sds}
%%%%%
Schema diagrams are intuitive visual representations of the structure of an ontology. They are, in general, \emph{not} unambiguous. In particular, this means that a particular node-edge-node construction can take on many different meanings. Instead, the idea is to communicate that there is some relation between the two concepts, and the more exact nature of the relation is deferred to the formal axiomatization which is expressed in OWL or some other suitable logic \cite{FOST}. For additional reading on the meaning behind schema diagram edges, see \cite{sdont,owlaxax}. For this paper, we have modified the traditional syntax used in the MOMo \cite{momo-swj} methodology to better communicate the important conceptual components when modeling with Wikibase, while acknowledging the underlying complexity of what Wikibase automatically generates during RDF serialization.

%%%%%%%%%%%%%%%%%%%%%%%%%%%%%%%%%%%%%%%%%%%%%%%%%%%%%%%%%%%%%%%%%
\subsection{The Graphical Syntax}
\label{ssec:syntax}
%%%%%
% Primary: Cogan, Andrew
% Explain the Graphical Syntax
%%%%%
In our portal, we provide additional variations of Figure~\ref{fig:pattern-generic} that are black \& white print friendly, and a color-blind palette.\footnote{\url{https://gitlab.cs.ksu.edu/daselab/wikibase-ontology-design-library}}

%% Expansions
\subsubsection*{Conceptual patterns and their expansions as diagrams}
In each diagram (Figures \ref{fig:pattern-generic}-\ref{fig:state-s}), we have paired a \emph{conceptual} diagram (the upper diagram) and its \emph{expansion}. The conceptual diagram, as we have previously described, is a method for graphically depicting, in a succinct manner, ``what is connected to what, and how?'' We have attempted to make this clear by pairing the label colors with border colors. That is, an orange label in the conceptual (top) diagram corresponds to the set of node-edge-node constructs with orange coloring or borders in the expanded (bottom) diagram.

%% Common Syntax
\subsubsection*{Common Diagram Syntax}
These colors and shapes are consistent across all diagrams in Figures~\ref{fig:pattern-generic}-\ref{fig:state-s}.
\begin{compactitem}
    \item Gold Rounded Rectangles: represent classes (objects).
    \item Purple Rounded Rectangles: represent classes belonging to the Wikibase namespace.
    \item Yellow Ellipses: represent datatypes. 
    \item Solid Arrow Heads: represent binary relations. If the target of the arrow is an ellipse, then it is a data property. Otherwise it is an object property.
    \item Open Arrow Heads: represent a \emph{subclass} relation.
    \item Dashed Edge Lines: represent an \emph{instanceOf} relation.
\end{compactitem}
%% Condensed Syntax
\subsubsection*{Condensed Diagram Syntax}
These colors and shapes are for the upper diagrams in Figures~\ref{fig:pattern-generic}-\ref{fig:state-s}.
\begin{compactitem}
    \item Green Rectangles: correspond to a \emph{collapsed} statement. It hides the underlying \textsf{wikibase:Statement} node and connecting properties.
    \item Orange Rectangles: correspond to a \emph{collapsed} qualifier.
    \item Diamond Arrow Heads: represent a connection to a qualifier.
    \item Blue Rectangles: correspond to a \emph{collapsed} reference. 
    \item Circle Arrow Heads: represent a connection to a reference.
\end{compactitem}
%% Expanded Syntax
\subsubsection*{Expanded Diagram Syntax}
\begin{figure}[tb]
    \centering
    \includegraphics[width=\columnwidth]{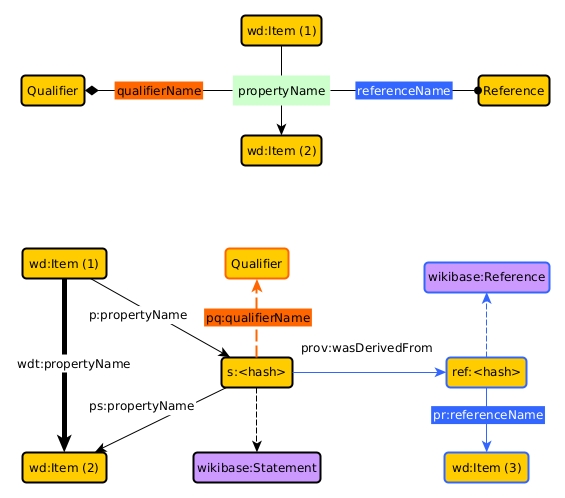}
    \caption{The top diagram shows a condensed view of the Wikibase conceptual model. The bottom diagram shows how the \emph{expansion} of the diagram to include the nodes that are automatically generated by the Wikibase framework. We use a diamond arrowhead and orange color to denote \textsf{Qualifier} nodes. A circle arrowhead and blue color to denote \textsf{Reference} nodes. In the expanded diagrams (in the lower portion), we change the border color of the nodes to indicate the origin of the generated nodes.}
    \label{fig:pattern-generic}
\end{figure}

\begin{figure}[tb]
    \centering
    \includegraphics[width=\columnwidth]{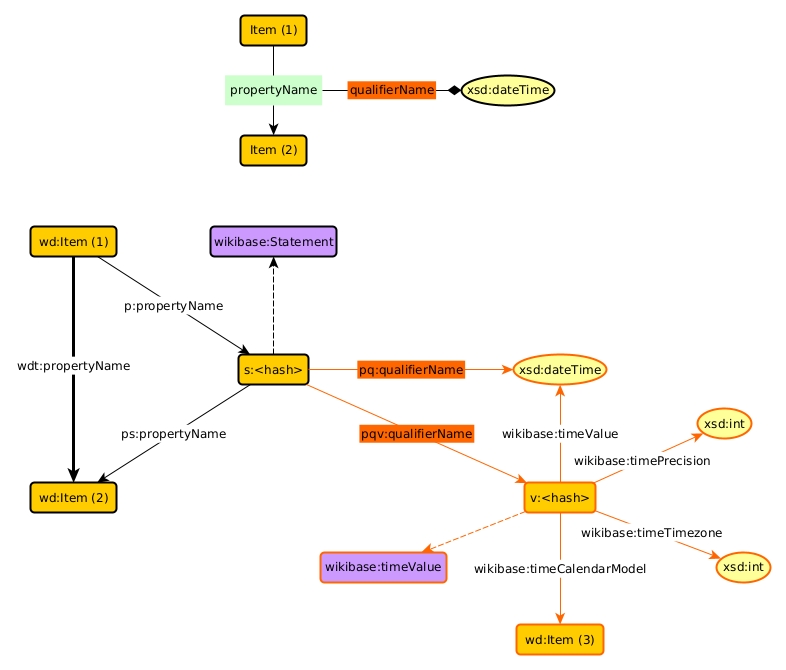}
    \caption{Condensed (top) and expanded (bottom) schema diagrams with an \textsf{xsd:datetime} for a qualifier. Nodes bordered in orange expand from the label node in the top diagram.}
    \label{fig:qual-dt}
\end{figure}

\begin{figure}[tb]
    \centering
    \includegraphics[width=\columnwidth]{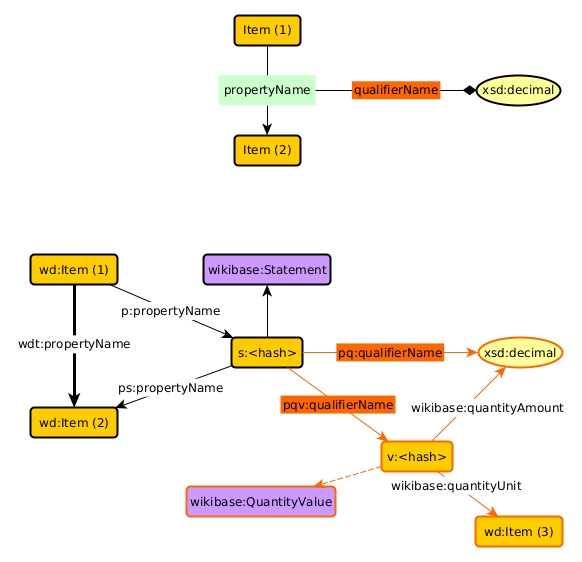}
    \caption{Condensed (top) and expanded (bottom) schema diagrams for modeling with \textsf{xsd:decimal} as a qualifier to \textsf{wdt:propertyName}.}
    \label{fig:qual-q}
\end{figure}

\begin{figure}[tb]
    \centering
    \includegraphics[width=\columnwidth]{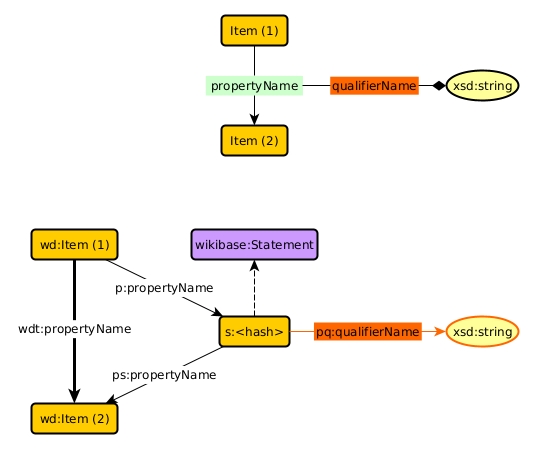}
    \caption{Condensed (top) and expanded (bottom) schema diagrams for modeling with \textsf{xsd:string} as a qualifier for \textsf{wdt:propertyName}.}
    \label{fig:qual-s}
\end{figure}

\begin{figure}[tb]
    \centering
    \includegraphics[width=\columnwidth]{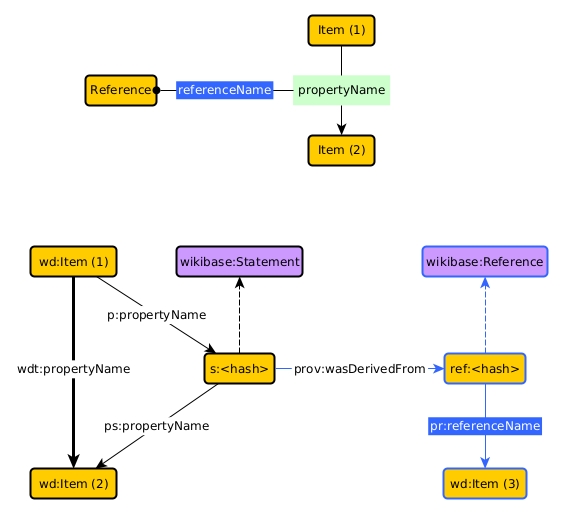}
    \caption{Condensed (top) and expanded (bottom) schema diagrams including a Reference for the assertions made by \textsf{wdt:propertyName}.}
    \label{fig:ref-nq}
\end{figure}

\begin{figure}[tb]
    \centering
    \includegraphics[width=\columnwidth]{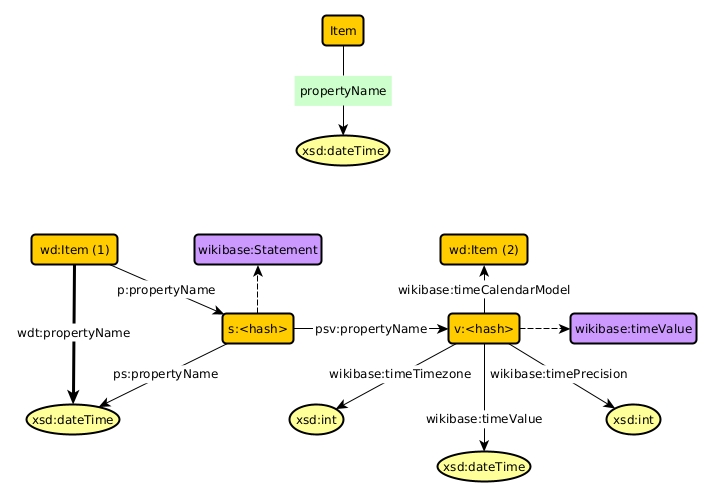}
    \caption{Condensed (top) and expanded (bottom) schema diagrams when modeling \textsf{wdt:propertyName} as a data property and the datatype is a \textsf{xsd:datetime}.}
    \label{fig:state-dt}
\end{figure}

\begin{figure}[tb]
    \centering
    \includegraphics[width=\columnwidth]{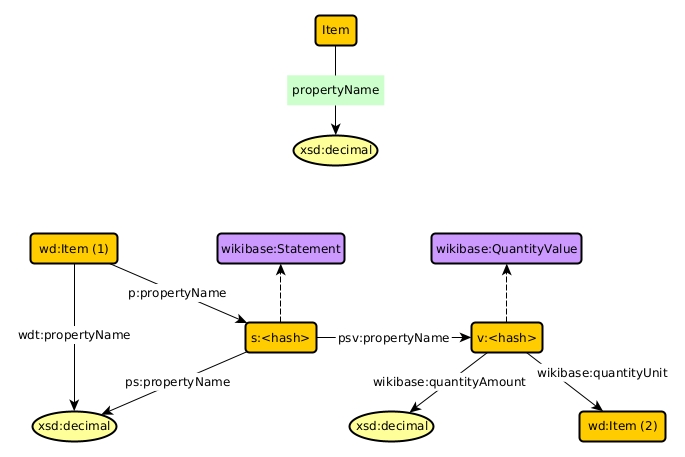}
    \caption{Condensed (top) and expanded (bottom) schema diagrams when modeling \textsf{wdt:propertyName} as a data property and the datatype is a \textsf{xsd:decimal}.}
    \label{fig:state-q}
\end{figure}

These colors and shapes are for the lower diagrams in Figures~\ref{fig:pattern-generic}-\ref{fig:state-s}.
\begin{compactitem}
    \item Orange Colored Borders: indicate that the edge or rectangle was originally hidden by the corresponding collapsed label in the above diagram. Orange indicates that the collapsed label is regarded a Qualifier.
    \item Blue Colored Borders: indicate that the edge or rectangle was originally hidden by the corresponding collapsed label in the above diagram. Blue indicates that the collapsed label is regarded a Reference.
    \item Hash Nodes: represent instances which are automatically generated according to some hashing algorithm. In these diagrams there are two appearances, in the \textsf{s:} namespace, for \textsf{wikibase:Statement}s, and the \textsf{ref:} namespace, for References on \textsf{wikibase:Statement}.
\end{compactitem}

%%%%%%%%%%%%%%%%%%%%%%%%%%%%%%%%%%%%%%%%%%%%%%%%%%%%%%%%%%%%%%%%%
\subsection{Axiomatization}
\label{ssec:axioms}
%%%%%
% What is the purpose of the axiomatization?
The purpose of this section is to provide a reference axiomatization of the Wikibase model. This allows us to specify axioms over the conceptual diagrams, which we detail in Section~\ref{sec:con-mod}. These axioms are provided in Description Logic, for a primer on this and the notation, please see \cite{FOST}.
Recall that \textsf{propertyName}, \textsf{qualifierName}, and \textsf{referenceName} are \emph{placeholders} to improve the clarity of these diagrams. They are meant to be replaced when utilizing these patterns and their expansions. Indeed, when replacing one, the corresponding occurrences will all be replaced by the same predicate name, and they will then be distinguishable by their namespaces.

\begin{figure}[tb]
    \centering
    \includegraphics[width=.7\columnwidth]{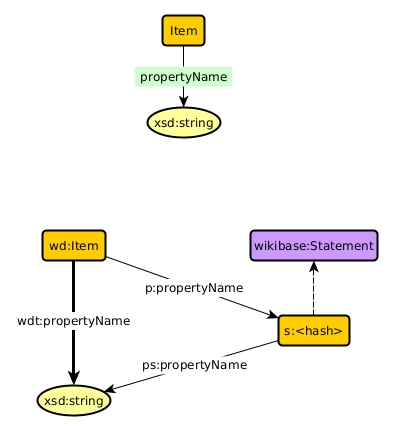}
    \caption{Condensed (top) and expanded (bottom) schema diagrams when modeling \textsf{wdt:propertyName} as a data property and the datatype is a \textsf{xsd:string}.}
    \label{fig:state-s}
\end{figure}

For example, we may replace \textsf{wdt:propertyName} with \textsf{wdt:hasName} and have the edge point to an \textsf{xsd:string}. At that point, we would replace all instances of ``propertyName'' with ``hasName'' across all namespaces in the diagram. We would then use Figure~\ref{fig:state-s} and the accompanying axiomatization. Additional examples can be found in Section~\ref{sec:example}. Note that we have used the \textsf{xsd:} namespace, which stands for XML Schema Datatypes. These are a W3C standard way of representing data primitives \cite{xsd-tr}.

Finally, recall that many of these structures can be directly seen in Figure \ref{fig:wb-rdf}.

% What's in this section?
In the following sections, we provide the description logic formulation of the axioms.\footnote{Additional information on the syntax and construction can be found in \cite{FOST,dl-primer}.}
% Shortened Prefixes
We have shortened some prefixes and predicates due to length: 
\begin{compactenum}
    \item \textsf{wikibase:} has been shortened to \textsf{wb:}
    \item \textsf{propertyName} has been shortened to \textsf{propName}
    \item \textsf{qualifierName} has been shortened to \textsf{qualName}
    \item \textsf{referenceName} has been shortened to \textsf{refName}
    \item \textsf{wasDerivedFrom} has been shortened to \textsf{wDF}
\end{compactenum}
%%%%%%%%%%%%%%%%%%%%%%%%%%%%%%%%%%%%%%%%%%%%%%%%%%%%%%%%%%%%%%%%%
%% Figures
%%%%%
\begin{figure}[tb]
    \centering
    \includegraphics[width=\columnwidth]{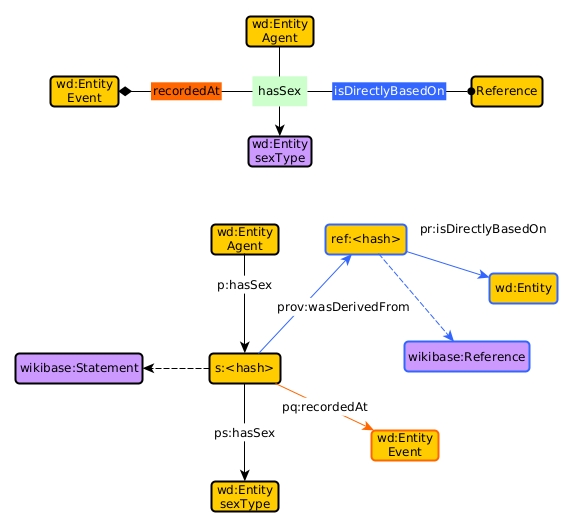}
    \caption{The reconstructed module for \textsf{SexRecord} from the Enslaved Ontology now using the Wikibase patterns.}
    \label{fig:sex-rec}
\end{figure}
\begin{figure}[tb]
    \centering
    \includegraphics[width=\columnwidth]{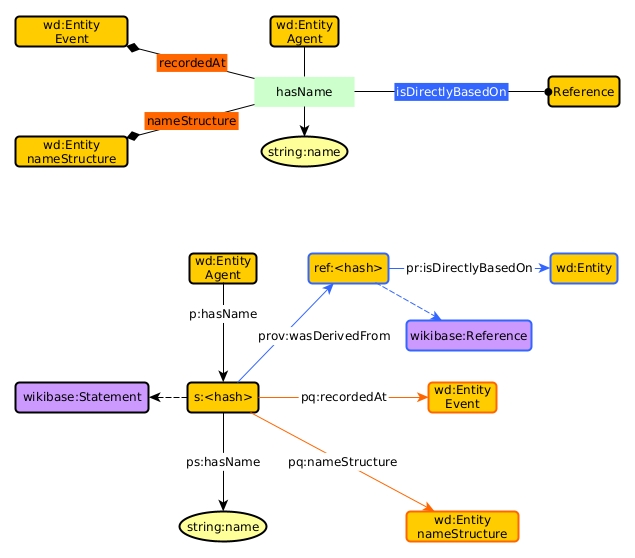}
    \caption{The reconstructed module for \textsf{NameRecord} from the Enslaved Ontology now using the Wikibase patterns.}
    \label{fig:name-rec}
\end{figure}
\begin{figure}[tb]
    \centering
    \includegraphics[width=\columnwidth]{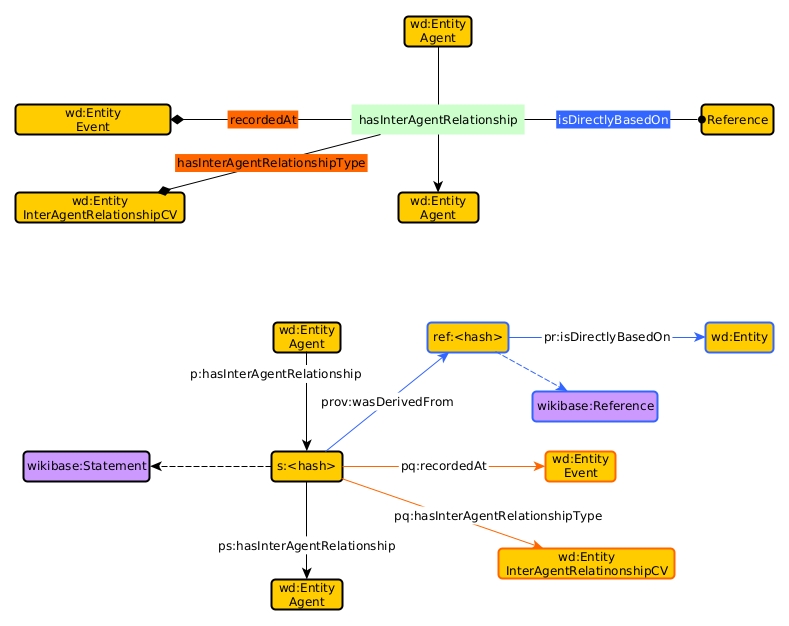}
    \caption{The reconstructed module for \textsf{InteragentRelationshipRecord} from the Enslaved Ontology now using the Wikibase patterns.}
    \label{fig:iar-rec}
\end{figure}
\begin{figure}[tb]
    \centering
    \includegraphics[width=\columnwidth]{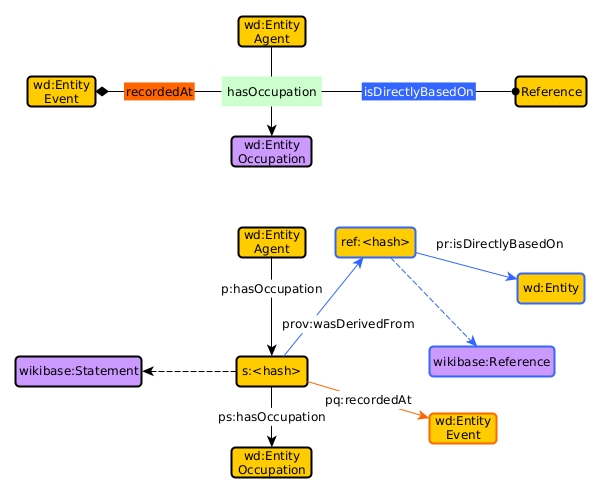}
    \caption{The reconstructed module for \textsf{OccupationRecord} from the Enslaved Ontology now using the Wikibase patterns.}
    \label{fig:occ-rec}
\end{figure}
\begin{figure}[tb]
    \centering
    \includegraphics[width=\columnwidth]{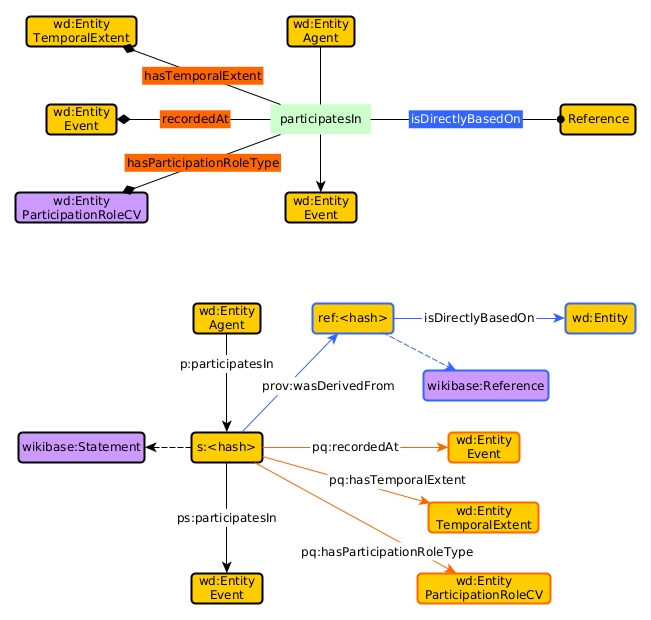}
    \caption{The reconstructed module for \textsf{ParticipatesInRecord} from the Enslaved Ontology now using the Wikibase patterns.}
    \label{fig:part-rec}
\end{figure}
\begin{figure*}[tb]
    \centering
    \includegraphics[width=.8\textwidth]{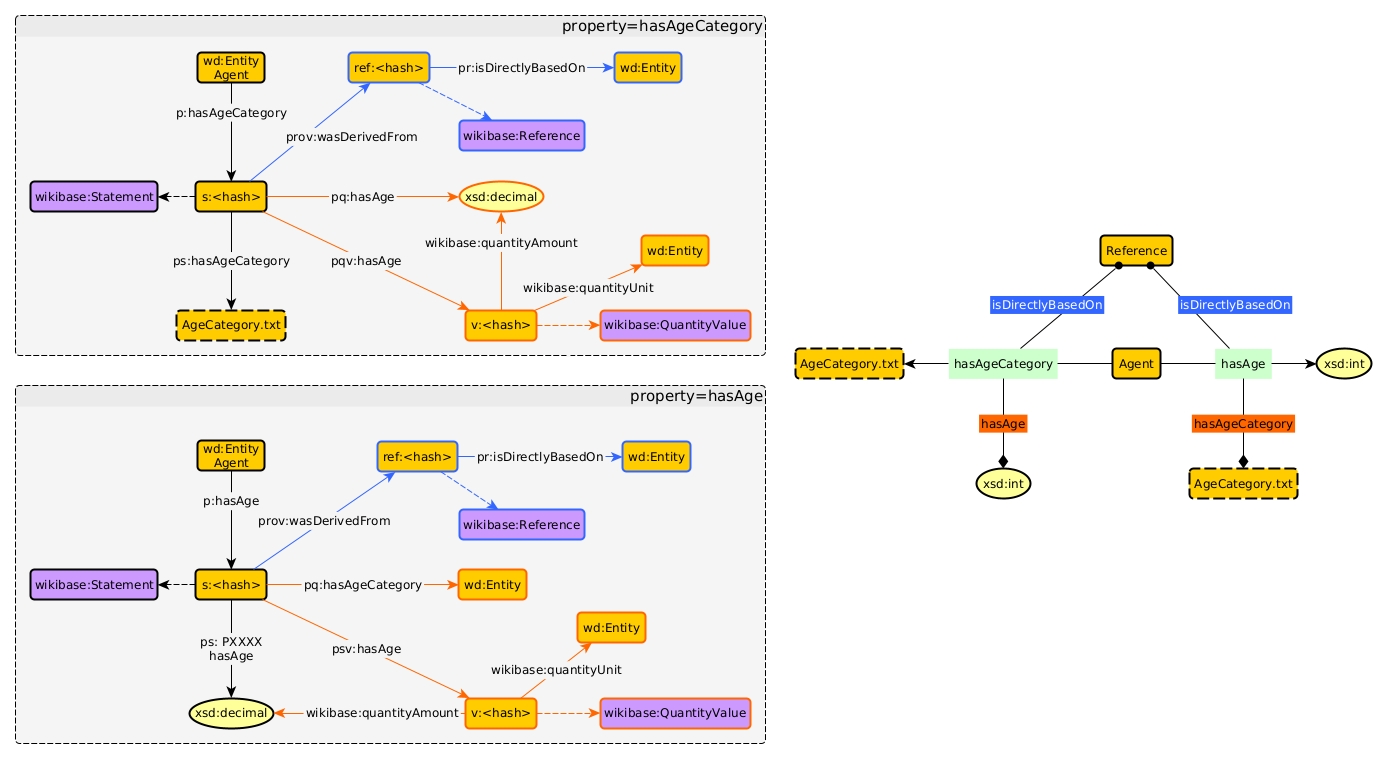}
    \caption{The reconstructed module for \textsf{AgeRecord} from the Enslaved Ontology now using the Wikibase patterns.}
    \label{fig:age-rec}
\end{figure*}
%%%%%
%%%%%%%%%%%%%%%%%%%%%%%%%%%%%%%%%%%%%%%%%%%%%%%%%%%%%%%%%%%%%%%%%
\subsubsection{Axioms Invariant to Exapnsion Type}
\label{sssec:invariant}
%%%%%
In this section, we discuss those axioms that appear to be invariant under the expansion type, as seen in Figure~\ref{fig:pattern-generic}.
%%%%%
\begin{align}
    % (Axiom 1) p:propertyName domain -> ?
    \exists ~\textsf{p:propName}.\top &\sqsubseteq \textsf{wb:Item}
\end{align}
%%%%%
Axiom 1 restricts the domain of \textsf{p:propName} to \textsf{wb:Item}.
%%%%%
\begin{align}
    % (Axiom 2) p:propertyName range -> wikibase:Statement
    \top &\sqsubseteq \phantom{\mathord{\leq}}\forall~\textsf{p:propName.wb:Statement} \\
    % (Axiom 3) p:propertyName inverse functional
    \top &\sqsubseteq \mathord{\leq}1~\textsf{p:propName}^-\textsf{.wb:Statement} \\
    % (Axiom 4) p:propertyName inverse existential
    \top &\sqsubseteq \phantom{\mathord{\leq}}\exists~\textsf{p:propName}^-\textsf{.wb:Statement}
\end{align}
%%%%
Axiom 2 restricts the range of \textsf{p:propName} to \textsf{wb:Statement}. Recall that \textsf{wb:Statement} instances are reifications of the \textsf{wdt:propName} relation, and carry the qualifications and reference information about a particular triple (i.e., \emph{statements}).
Axiom 3 states that the inverse of \textsf{p:propName} is functional, (i.e., inverse functionality).
Axiom 4 states that the inverse of \textsf{p:propName} is existential (i.e., inverse existential). Together, Axioms 3 and 4 indicate that a \textsf{wb:Statement} has exactly one inverse filler for \textsf{p:propName$^-1$}. This may also be specified more succinctly using exact cardinality.
\begin{equation*}
    \top \sqsubseteq \mathord{=}1~\textsf{p:propName}^-\textsf{.wb:Statement}
\end{equation*}

%%%%
\begin{align}
    % (Axiom 5) ps:propertyName domain -> wikibase:Statement
    \exists ~\textsf{ps:propName}.\top &\sqsubseteq \textsf{wb:Statement}
\end{align}
%%%%
Axiom 5 restricts the domain of \textsf{ps:propName} to \textsf{wb:Statement}. Essentially, as a part of the reification of \textsf{wdt:propName}, \textsf{ps:propName} will always have an inverse filler of \textsf{wb:Statement}.

%%%%
\begin{align}
    % (Axiom 6) ps:propertyName range -> wikibase:Item
\top &\sqsubseteq \phantom{\mathord{\leq}}\forall~\textsf{p:propName.wb:Item} \\
    % (Axiom 7) ps:propertyName functional
    \top &\sqsubseteq \mathord{=}1~\textsf{ps:propName.wb:Item}
\end{align}
%%%%%
Axiom 6 restricts the range of \textsf{ps:propName} to \textsf{wb:Item}, acting as the other half of the reification for \textsf{wdt:propName}. 
Axiom 7 states that \textsf{ps:propName} has exactly one filler of type \textsf{wb:Item}, in the same manner as Axiom 4. These functionality and existential restrictions are necessary to mimic how and why these \textsf{wb:Statement} nodes are created. They  \emph{uniquely} connect the \textsf{wdt:propName} to the qualifier and reference information.

%%%%%
\begin{align}
    % (Axiom 8) pq:qualifierName domain -> wikibaseStatement
    \exists ~\textsf{pq:qualifierName}.\top &\sqsubseteq \textsf{wb:Statement}
\end{align}
%%%%%
Axiom 8 restricts the domain of any \textsf{pq:qualifierName} to \textsf{wb:Statement}. In the Wikibase model, qualifiers are always attached to the ``hash nodes.'' We will discuss in the later sections specific to the type of qualifier the axioms pertaining to range, functionality, and, etc.

%%%%%%%%%%%%%%%%%%%%%%%%%%%%%%%%%%%%%%%%%%%%%%%%%%%%%%%%%%%%%%%%%
\subsubsection{Axioms between Items}
\label{sssec:ax-item}
%%%%%
\begin{align}
    % (Axiom 9) p:propertyName o ps:propertyName $\sqsubseteq$ wdt:propertyName
    \textsf{p:propName} \circ \textsf{ps:propName} &\sqsubseteq \textsf{wdt:propName}
\end{align}
%%%%%
Axiom 9 is a role chain that formalizes the reification of \textsf{wdt:propName}. From the previous axioms, we know that the connecting filler for the reification is of type \textsf{wb:Statement}. Additionally, we can infer the following.
%%%%%
\begin{align*}
    % wdt:propertyName domain -> wb:Item (same as p:propertyName)
    \exists~\textsf{wdt:propName}.\top &\sqsubseteq \textsf{wb:Item} \\
    % wdt:propertyName scoped domain -> wb:Item (can be expressed over role chain?)
    \exists~\textsf{wdt:propName.wb:Item} &\sqsubseteq \textsf{wb:Item} \\
    % wdt:propertyName range -> wb:Item (same as ps:propertyName)
    \exists~\textsf{wdt:propName}^-.\top &\sqsubseteq \textsf{wb:Item} \\
    % wdt:propertyName scoped range -> wb:Item (can not be expressed over role chain?)
    \exists~\textsf{wdt:propName}^-.\textsf{wb:Item} &\sqsubseteq \textsf{wb:Item}
\end{align*}
%%%%%
Note that the latter two axioms are written using the inverse form, as we wish to avoid having a role chain appear on the right hand side of the axiom, which cannot be specified in OWL 2.

%%%%%%%%%%%%%%%%%%%%%%%%%%%%%%%%%%%%%%%%%%%%%%%%%%%%%%%%%%%%%%%%%
\subsubsection{Axioms Invariant for Qualifiers}
\label{sssec:ax-qual}
%%%%%
Each type of qualifier that Wikibase supports (i.e., Date, Numeric, and String) have different structures. In this section, we only discuss those axioms that are invariant across these expansions. For reference, they pertain to the expansions as seen in Figures~\ref{fig:qual-dt}-\ref{fig:qual-s}. 

There are two ways to restrict the nature of a qualifier (i.e., the type) for a specific \textsf{pq:qualName}: scoped and un-scoped. Scoped, in this case, means that the type of the Item being qualified matters, whereas \emph{un-scoped} is a global restriction on \textsf{pq:qualName}.

The scoped restriction follows, in rule format.
%%%%%
\begin{align*}
    \forall x,y,z ~\textsf{Item}(x)~&\wedge~\textsf{p:propName}(x,y)~\notag\\&\wedge~\textsf{pq:qualName}(y,z)~\longrightarrow~\textsf{Qualifier}(z)
\end{align*}
%%%%%
Note that in order to convert this into description logic, we will need to use a complex, inverse domain restriction here. Since a qualification occurs on a \textsf{wb:Statement} instance, we need to construct a role chain to the \textsf{wb:Item} and, as previously noted, we cannot have role chains on the right hand side of an axiom in OWL 2, we take the inverse of the range restriction, as seen in Axiom 10.
%%%%%
\begin{equation}
    % Axiom 10 (range restriction on type of qualifier for a type)
    \exists~\textsf{pq:qualName}^-.\left(\exists~\textsf{p:propName}^-\textsf{.Item}\right) \sqsubseteq \textsf{Qualifier}
\end{equation}
%%%%%
Specifying an un-scoped range restriction on \textsf{pq:qualName} is much simpler (Axiom 11).
%%%%%
\begin{equation}
    % Axiom 11 (require a qualifier of a certain type)
    \top \sqsubseteq \forall~\textsf{pq:qualName.Qualifier}
\end{equation}
%%%%%
We also note that the domain of \textsf{pq:qualName} is always a \textsf{wb:Statement}.
\begin{equation}
    \exists~\textsf{pq:qualName.}\top \sqsubseteq \textsf{wb:Statement}
\end{equation}

%%%%%%%%%%%%%%%%%%%%%%%%%%%%%%%%%%%%%%%%%%%%%%%%%%%%%%%%%%%%%%%%%
\subsubsection{Axioms for Date Qualifiers}
\label{sssec:ax-qual-dt}
%%%%%
This section covers the axioms pertaining to the Wikibase model when the qualifier is a datetime. A graphical view of this structure is shown in Figure~\ref{fig:qual-dt}. This expansion creates a construction that incorporates metadata for the datetime, such as the timezone and the temporal reference system, which is done by creating a hash node in the \textsf{v:} namespace.

% pq:qualName gd, gr/sr
Axiom 13 indicates that the domain of \textsf{pq:qualName} is always \textsf{wb:Statement}.
\begin{equation}
    \exists~\textsf{pq:qualName.}\top \sqsubseteq \textsf{wb:Statement}
\end{equation}
%%%%%
Axiom 14 is the specific version of Axiom 10 (scoped range restriction) for the \textsf{xsd:dateTime} qualifier.
\begin{equation}
    \exists~\textsf{pq:qualName}^-.\left(\exists~\textsf{p:propName}^-\textsf{.Item}\right) \sqsubseteq \textsf{xsd:dateTime}
\end{equation}
%%%%%
Axiom 15 is the specific version of Axiom 11 (un-scoped range restriction) for the \textsf{xsd:dateTime} qualifier. Based on modeling needs, one would choose only Axiom 14 or Axiom 15.
\begin{equation}
    \top \sqsubseteq \forall~\textsf{pq:qualName.wb:timeValue}
\end{equation}
%%%%%
% pqv:qualName gd, gr, functional, inverse functional and existential
Axiom 16 indicates that the domain of \textsf{pqv:qualName} is restricted to \textsf{wb:Statement}.
\begin{equation}
    \exists~\textsf{pqv:qualName.}\top \sqsubseteq \textsf{wb:TimeValue}
\end{equation}
%%%%%
Axiom 17 specifies that the range of \textsf{pqv:qualName} is restricted to \textsf{wb:TimeValue}.
\begin{align}
    \top &\sqsubseteq \forall~\textsf{pqv:qualName.wb:timeValue}
\end{align}
%%%%%
In Axiom 18, we formalize the notion that \textsf{wb:TimeValue} instances are unique. That is, they are the filler for at most one \textsf{pqv:qualName} triple, and that it must qualify a \textsf{wb:Statement}.
\begin{equation}
    \textsf{wb:TimeValue} \sqsubseteq \mathord{=}1~\textsf{pqv:qualName}^-\textsf{.wb:Statement}
\end{equation}
%%%%%
We specify that the domain for each of the predicates (in the \textsf{wikibase} namespace, \textsf{timeValue}, \textsf{timePrecision}, \textsf{timeTimezone} and \textsf{timeCalendarModel},is \textsf{wb:timeValue} (which appears as a hashed node in the \textsf{v:} namespace in Figure \ref{fig:qual-dt}). This is specified in Axioms 19-22.
\begin{align}
    \exists~\textsf{timeValue.}\top &\sqsubseteq \textsf{wb:timeValue} \\
    \exists~\textsf{timePrecision.}\top &\sqsubseteq \textsf{wb:timeValue} \\
    \exists~\textsf{timeTimezone.}\top &\sqsubseteq \textsf{wb:timeValue} \\
    \exists~\textsf{timeCalendarModel.}\top &\sqsubseteq \textsf{wb:timeValue} 
\end{align}

We may also specify their ranges, globally, in Axioms 23-26. We also indicate that they have an exact cardinality of one in Axioms 27-30.
\begin{align}
	\top &\sqsubseteq \forall~\textsf{wb:timeValue.xsd:dateTime} \\
	\top &\sqsubseteq \forall~\textsf{wb:timePrecision.xsd:int} \\
	\top &\sqsubseteq \forall~\textsf{wb:timeTimezone.xsd:int} \\
	\top &\sqsubseteq \forall~\textsf{wb:timeCalendarModel.wd:Item} \\
    \top &\sqsubseteq \mathord{=}1~\textsf{wb:timeValue.xsd:dateTime} \\
    \top &\sqsubseteq \mathord{=}1~\textsf{wb:timePrecision.xsd:int} \\
    \top &\sqsubseteq \mathord{=}1~\textsf{wb:timeTimezone.xsd:int} \\
    \top &\sqsubseteq \mathord{=}1~\textsf{wb:timeCalendarModel.wd:Item}
\end{align}
Finally, it should be noted that whenever an assertion of \textsf{pq:qualName} exists, there is an associated ``hash node'' that accompanies it. We specify this in Axiom 31. 
\begin{align}
    \exists~\textsf{pq:qualName.}&\textsf{xsd:dateTime} \sqsubseteq \nonumber\\ &\exists~\textsf{pqv:qualName.wb:TimeValue}
\end{align}
This ensures that the node carrying the additional contextual data is connected to the interesting property name, and connected back to the statement.
%%%%%%%%%%%%%%%%%%%%%%%%%%%%%%%%%%%%%%%%%%%%%%%%%%%%%%%%%%%%%%%%%
\subsubsection{Axioms for String Qualifiers}
\label{sssec:ax-qual-s}
%%%%%
In this section, we cover the axioms pertaining to the Wikibase model when the qualifier is a string. A graphical view of this structure is shown in Figure~\ref{fig:qual-s}.
% pq:qualName gd, gr/sr
Axiom 32 indicates that the domain of \textsf{pq:qualName} is always \textsf{wb:Statement}.
\begin{equation}
    \exists~\textsf{pq:qualName.}\top \sqsubseteq \textsf{wb:Statement}
\end{equation}
%%%%%
Axiom 33 is the specific version of Axiom 10 (scoped range restriction) for the \textsf{xsd:dateTime} qualifier.
\begin{equation}
    \exists~\textsf{pq:qualName}^-.\left(\exists~\textsf{p:propName}^-\textsf{.Item}\right) \sqsubseteq \textsf{xsd:string}
\end{equation}
%%%%%
Axiom 34 is the specific version of Axiom 11 (un-scoped range restriction) for the \textsf{xsd:dateTime} qualifier. Based on modeling needs, one would choose only Axiom 33 or Axiom 34.
\begin{equation}
    \top \sqsubseteq \forall~\textsf{pq:qualName.xsd:string}
\end{equation}

%%%%%%%%%%%%%%%%%%%%%%%%%%%%%%%%%%%%%%%%%%%%%%%%%%%%%%%%%%%%%%%%%
\subsubsection{Axioms for Numeric Value Qualifiers}
\label{sssec:ax-qual-q}
%%%%%
Wikibase uses \textsf{xsd:decimal} with \textsf{wikibase:QuantityUnit} for this purpose. The pattern for using \textsf{xsd:decimal} as a qualifier can be found in Figure~\ref{fig:qual-q}. Qualifying with a quantity creates a construction that incorporates the unit of the quantity, which is also a \textsf{wb:Item}. It does this by creating another hash node of type \textsf{wb:QuantityValue}, this time in the \textsf{v:} namespace. The hash node is also referenced with \textsf{qualName}, but in the \textsf{pqv:} namespace. \textsf{pq:qualName} points directly to the amount.

% pq:qualName
\begin{align}
  % domain restriction
  \exists~\textsf{pq:qualName.}\top &\sqsubseteq \textsf{wb:Statement}  \label{ax:qn-d}\\
  % scoped range restriction
  \textsf{wb:Statement} &\sqsubseteq \forall~\textsf{pq:qualName.xsd:decimal}  \label{ax:qn-sr}\\
  % functional
  \textsf{wb:Statement} &\sqsubseteq \mathord{\leq}1~\textsf{pq:qualName.xsd:decimal}  \label{ax:qn-f}
\end{align}
% Axiom
Axiom \ref{ax:qn-d} is a domain restriction. That is, the inverse filler (i.e., the domain) is restricted to \textsf{wb:Statement}.
% Axiom
Axiom \ref{ax:qn-sr} is a scoped range restriction. That is, when the inverse filler (i.e., the domain) is of type \textsf{wb:Statement}, the filler (i.e., the range) of \textsf{pq:qualName} is restricted to \textsf{xsd:decimal}.
% Axiom
Axiom \ref{ax:qn-f} states that \textsf{pq:qualName} is functional. That is, a particular \textsf{wb:Statement} node targets at most one \textsf{xsd:decimal} via \textsf{pq:qualName}.
% pqv:qualName
\begin{align}
  % domain restriction
  \exists~\textsf{pqv:qualName.}\top &\sqsubseteq \textsf{wb:Statement}  \label{ax:qn-v-d}
\end{align}
% Axiom
Axiom \ref{ax:qn-v-d} is a domain restriction. That is, the inverse filler (i.e., the domain) is restricted to \textsf{wb:Statement}.
\begin{align}
  % scoped range restriction
  \textsf{wb:Statement} &\sqsubseteq \forall~\textsf{pqv:qualName.wb:QuantityValue}  \label{ax:qn-v-sr}\\
  % functional
  \textsf{wb:Statement} &\sqsubseteq \mathord{\leq}1~\textsf{pqv:qualName.wb:QuantityValue}  \label{ax:qn-v-f}
\end{align}
% Axiom
Axiom \ref{ax:qn-v-sr} is a scoped range restriction. That is, when the inverse filler (i.e., the domain) is of type \textsf{wb:Statement}, the filler (i.e., the range) of \textsf{pqv:qualName} is restricted to \textsf{wb:QuantityValue}.
% Axiom
Axiom \ref{ax:qn-v-f} states that \textsf{pqv:qualName} is functional. That is, a particular \textsf{wb:Statement} node targets at most one \textsf{wb:QuantityValue} via \textsf{pqv:qualName}.

% wb:quantityValue
\begin{align}
  % domain restriction
  \exists~\textsf{wb:quantityValue.}\top &\sqsubseteq \textsf{wb:QuantityValue}  \label{ax:qv-d}
\end{align}
\begin{align}
  % scoped range restriction
  \textsf{wb:QuantityValue} &\sqsubseteq \forall~\textsf{wb:quantityValue.xsd:decimal}  \label{ax:qv-sr}\\
  % existential
  \textsf{wb:QuantityValue} &\sqsubseteq \exists~\textsf{wb:quantityValue.wb:QuantityValue}  \label{ax:qv-e}\\
  % functional
  \textsf{wb:QuantityValue} &\sqsubseteq \mathord{\leq}1~\textsf{wb:quantityValue.xsd:decimal}  \label{ax:qv-f}
\end{align}
% Axiom
Axiom \ref{ax:qv-d} is a domain restriction. That is, the inverse filler (i.e., the domain) is restricted to \textsf{wb:QuantityValue}.
% Axiom
Axiom \ref{ax:qv-sr} is a scoped range restriction. That is, when the inverse filler (i.e., the domain) is of type \textsf{wb:QuantityValue}, the filler (i.e., the range) of \textsf{wb:quantityValue} is restricted to \textsf{xsd:decimal}.
% Axiom
Axiom \ref{ax:qv-e} states that \textsf{wb:quantityValue} is existential. That is, there is always at least one filler.
% Axiom
Axiom \ref{ax:qv-f} states that \textsf{wb:quantityValue} is functional. That is, a particular \textsf{wb:QuantityValue} node targets at most one \textsf{xsd:decimal} via \textsf{wb:quantityValue}.

% wb:quantityUnit
\begin{align}
  % domain restriction
  \exists~\textsf{wb:quantityUnit.}\top &\sqsubseteq \textsf{wb:QuantityValue}  \label{ax:qu-d}\\
  % scoped range restriction
  \textsf{wb:QuantityValue} &\sqsubseteq \forall~\textsf{wb:quantityUnit.wb:Item}  \label{ax:qu-sr}\\
  % existential
  \textsf{wb:QuantityValue} &\sqsubseteq \exists~\textsf{wb:quantityUnit.wb:QuantityValue}  \label{ax:qu-e}\\
  % functional
  \textsf{wb:QuantityValue} &\sqsubseteq \mathord{\leq}1~\textsf{wb:quantityUnit.wb:Item}  \label{ax:qu-f}
\end{align}
% Axiom
Axiom \ref{ax:qu-d} is a domain restriction. That is, the inverse filler (i.e., the domain) is restricted to \textsf{wb:QuantityValue}.
% Axiom
Axiom \ref{ax:qu-sr} is a scoped range restriction. That is, when the inverse filler (i.e., the domain) is of type \textsf{wb:QuantityValue}, the filler (i.e., the range) of \textsf{wb:quantityUnit} is restricted to \textsf{wb:Item}.
% Axiom
Axiom \ref{ax:qu-e} states that \textsf{wb:quantityUnit} is existential. That is, there is always least one filler.
% Axiom
Axiom \ref{ax:qu-f} states that \textsf{wb:quantityUnit} is functional. That is, a particular \textsf{wb:QuantityValue} node targets at most one \textsf{wb:Item} via \textsf{wb:quantityUnit}.

%%%%%%%%%%%%%%%%%%%%%%%%%%%%%%%%%%%%%%%%%%%%%%%%%%%%%%%%%%%%%%%%%
\subsubsection{Axioms for Statements}
\label{sssec:ax-stat}
%%%%%
% pq:qualName gr/sr
Axiom 13 is the specific version of Axiom 10 (scoped range restriction) for the \textsf{xsd:dateTime} qualifier.
\begin{equation}
    \exists~\textsf{pq:qualName}^-.\left(\exists~\textsf{p:propName}^-\textsf{.Item}\right) \sqsubseteq \textsf{xsd:dateTime}
\end{equation}
%%%%%
Axiom 14 is the specific version of Axiom 11 (un-scoped range restriction) for the \textsf{xsd:dateTime} qualifier. Based on modeling needs, one would choose only Axiom 13 or Axiom 14.
\begin{equation}
    \top \sqsubseteq \forall~\textsf{pq:qualName.wb:timeValue}
\end{equation}
%%%%%

%%%%%%%%%%%%%%%%%%%%%%%%%%%%%%%%%%%%%%%%%%%%%%%%%%%%%%%%%%%%%%%%%
\subsubsection{Axioms for References}
\label{sssec:ax-ref}
%%%%%
In this section,

%%%%%
We only specify scoped range and domain restrictions over \textsf{prov:wasDerivedFrom}, in order to avoid unwanted ontological commitments in the event that the PROV Ontology \cite{provo} is used elsewhere in a developed ontology, as in Axioms \ref{ax:ref-wdf-sd} and \ref{ax:ref-wdf-sr}.
\begin{align}
    % (Axiom 10) prov:wasDerivedFrom domain -> wikibase:Statement
    \exists ~\textsf{prov:wDF.wb:Reference} &\sqsubseteq \textsf{wb:Statement} \label{ax:ref-wdf-sd}\\
    % (Axiom 11) prov:wasDerivedFrom range -> wikibase:Reference
    \textsf{wb:Statement} &\sqsubseteq \forall~\textsf{prov:wDF.wb:Reference} \label{ax:ref-wdf-sr}
\end{align}
%%%%%
It is always the case that the domain of \textsf{pr:refName} is a \textsf{wb:Reference} (Axiom \ref{ax:ref-rn-gd}).
\begin{equation}
    % Global Domain
    \exists~\textsf{pr:refName.}\top \sqsubseteq \textsf{wb:Reference} \label{ax:ref-rn-gd}
\end{equation}
%%%%
In the same manner that we have specified the range of \textsf{pq:qualName} (i.e. Axioms 10 and 11), we will for \textsf{pr:referenceName}.
\begin{align}
    % Scoped Range
    &\exists~\textsf{pr:referenceName}^-.\nonumber\\&\quad(\exists~\textsf{prov:wDF}^-.(\exists~\textsf{p:propName}^-.\top))\nonumber\\&\qquad\sqsubseteq \textsf{wb:Item} \label{ax:ref-rn-sr}
\end{align}
\begin{equation}
    % Global Range
    \exists ~\textsf{pr:refName.}\top \sqsubseteq \textsf{wb:Reference} \label{ax:ref-rn-gr}
\end{equation}

Furthermore, if we wish to vary the type of reference based on the domain of the statement assertion, we will need to construct an axiom similar to Axiom \ref{ax:ref-rn-sr}.
\begin{align}
    % Scoped Domain
    &\exists~\textsf{p:propName.}\nonumber\\&\quad\left(\exists~\textsf{prov:wDF.}\left(\exists~\textsf{pr:refName.}\top\right)\right)\nonumber\\&\qquad\sqsubseteq \textsf{wb:Item} \label{ax:ref-rn-sd}
\end{align}
Finally, we specify that a particular reference node is uniquely the target of a Statement node. That is, \textsf{prov:wasDerivedFrom} is restricted by an inverse existential and inverse functionality axiom, which together form Axiom \ref{ax:ref-wdf-ec}.
\begin{equation}
    % Inverse exact cardinality
    \textsf{wb:Reference} \sqsubseteq \mathord{=}1~\textsf{prov:wDF}^-.\textsf{wb:Reference} \label{ax:ref-wdf-ec}
\end{equation}
%%%%%

%%%%%%%%%%%%%%%%%%%%%%%%%%%%%%%%%%%%%%%%%%%%%%%%%%%%%%%%%%%%%%%%%
\subsection{Resources}
\label{ssec:resources}
%%%%%
We have included a series of data shapes, for the purposes of validating triples materialized against the axioms described above. These are expressed in ShEx, the Shape Expression Language\footnote{\url{https://shex.io/}}, which is a structural schema language for RDF graphs. These are available online. These are expressed in ShEx as that seems to be the Wikidata community's adopted way of validating data.

We have also provided two serializations for this manuscript: an OWL file containing the axiomatization of the Wikibase conceptual model and an OWL file containing the list of axiom patterns, as described in Section \ref{sec:con-mod}. 

All of these resources are available online under Apache License 2.0.\footnote{See \url{https://gitlab.cs.ksu.edu/daselab/wikibase-ontology-design-library}. We will provide a persistent URI in the final version of the manuscript.}

%%%%%%%%%%%%%%%%%%%%%%%%%%%%%%%%%%%%%%%%%%%%%%%%%%%%%%%%%%%%%%%%
\section{Conceptual Modeling}
\label{sec:con-mod}
%%%%%
% The overarching purpose of this axiomatization is to be able to conceptually reason about the Wikibase statements, without worrying about the underlying Wikibase structure. 
As previously discussed, the overarching purpose of these patterns, and in particular the revised graphical syntax, is to simplify the discussion surrounding the Wikibase model. The next step is to improve our ability to conceptually reason about these patterns. That is, to be able to specify constraints and restrictions, such as mandating that statements of a certain type must always have a qualifier, again of a certain type. 

As such, we have taken the axiom patterns from \cite{owlaxax} and, for each such axiom pattern, provided a natural language approximation alongside the axiom patterns modified to suit the Wikibase model. In this way, we can utilize the natural language to do top-level conceptual reasoning and then, when it is time to formalize the model, map these simple, natural language statements into the formal axioms. Our methodology is as follows.

Recall that a triple has the form ``Subject Predicate Object''. One frequently encounters this via assertional statements, such as ``\textsf{ex:dog1 ex:hasName ``Fido''\^{}\^{}xsd:string}''. However, for the purposes of the following discussion, we will use Subject to refer, instead, to the Subject's Type (i.e., moving up a layer of abstraction). In this way, we would say ``ex:Dog ex:hasName xsd:string'', as it would appear as a node-edge-node construction in a schema diagram. In our axioms, we will use \textsf{ex:Sub}, \textsf{ex:Pred}, and \textsf{ex:Obj}, respectively.

In our natural language, we describe statements relative to the statement, and in particular the \textsf{ex:Pred} Statement. We will use \emph{about} to denote the \textsf{Subject} of a Statement and \emph{refers to} to denote the \textsf{Object} of the statement. For example, ``A \textsf{ex:hasName} Statement is always \emph{refers to} a \textsf{ex:Name},'' which we will see below as the natural language approximation of the Range axiom pattern.

We anticipate that these natural language approximations will be used in conjunction with the formal model included above in Section \ref{sec:con-mod}. As such, there may be some logical redundancies when the formal model is combined with the suggested axioms below. We do not consider this to be problematic and, indeed, believe that the inclusion of both can aid human understanding.

% Some of these axioms are also inferred due to the axioms global to the Wikibase structure, namely
Recall that
\begin{align}
    \textsf{p:propName} \circ \textsf{ps:propName} &\sqsubseteq \textsf{wdt:propName} \label{ax:role-chain}
\end{align}

\noindent\textbf{Domain:} ``A Predicate Statement is always about a Subject.''
\begin{align}
  \exists \textsf{p:propName.}\top   &\sqsubseteq \textsf{ex:Sub} \\
  \exists \textsf{wdt:propName.}\top &\sqsubseteq \textsf{ex:Sub}
\end{align}

\noindent\textbf{Range:} ``A Predicate Statement always refers to an Object.''
\begin{align}
  \top &\sqsubseteq \forall \textsf{ps:propName.ex:Obj} \\
  \top &\sqsubseteq \forall \textsf{wdt:propName.ex:Obj}
\end{align}

\noindent\textbf{Scoped Domain:} ``A Predicate Statement that refers to an Object, is always about a Subject.''
\begin{align}
  \exists \textsf{p:propName.(}\exists\textsf{ps:propName.ex:Obj)} &\sqsubseteq \textsf{ex:Sub} \label{ax:cm-rc-sd}\\
  \exists \textsf{wdt:propName.ex:Obj} &\sqsubseteq \textsf{ex:Sub} \label{ax:cm-wpn-sd}
\end{align}
Axiom \ref{ax:cm-rc-sd} is one such logically redundant axiom: Axioms 5, 7, and \ref{ax:role-chain} together infer it. This fact is actually quite useful, as it means we may specify restrictions on \textsf{wdt:propName} and it remains consistent over the formal model.

% Scoped Range
\noindent\textbf{Scoped Range:} ``A Predicate Statement that is about a Subject always refers to an Object.''
\begin{align}
%   \textsf{ex:Sub} &\sqsubseteq \forall \textsf{ps:propName.ex:Obj} \\
  \textsf{ex:Sub} &\sqsubseteq \forall \textsf{wdt:propName.ex:Obj}
\end{align}
In line with Axiom 63, we should also have an axiom
\begin{equation*}
    \textsf{ex:Sub} \sqsubseteq \forall \textsf{(p:propName}\circ\textsf{ps:propName).ex:Obj}
\end{equation*}
but, it cannot be expressed directly in OWL. However, with Axioms 58 and 65, we can infer it.

% Functionality
\noindent\textbf{Functionality:} ``A Predicate Statement refers to at most one Item.''
\begin{align}
  \top &\sqsubseteq \mathord{\leq}1 \textsf{p:propName.}\top \\
  \top &\sqsubseteq \mathord{\leq}1 \textsf{wdt:propName.}\top
\end{align}
Note that there are variations of functionality, which we would call \emph{qualified} and \emph{scoped} based on whether or not the Predicate Statement is always about certain Subjects or refers to only certain Objects; these are described below.

% Inverse Functionality
\noindent\textbf{Inverse Functionality:} ``A Predicate Statement is about at most one Item.''
\begin{align}
  \top &\sqsubseteq \mathord{\leq}1 \textsf{ps:propName-.}\top \\
  \top &\sqsubseteq \mathord{\leq}1 \textsf{wdt:propName-.}\top
\end{align}

% Scoped Functionality
\noindent\textbf{Scoped Functionality:} ``A Predicate Statement is about at most one Subject.''
\begin{align}
  \textsf{ex:Sub} &\sqsubseteq \mathord{\leq}1 \textsf{p:propName.}\top \\
  \textsf{ex:Sub} &\sqsubseteq \mathord{\leq}1 \textsf{wdt:propName.}\top
\end{align}

% Qualified Functionality
\noindent\textbf{Qualified Functionality:} ``A Predicate Statement refers to at most one Object.''
\begin{align}
  \top &\sqsubseteq \mathord{\leq}1 \textsf{p:propName.}\top \\
  \top &\sqsubseteq \mathord{\leq}1 \textsf{wdt:propName.ex:Obj} \\
  \top &\sqsubseteq \mathord{\leq}1 \textsf{ps:propName.ex:Obj}
\end{align}
We need to include the third axiom which scopes the global functionality statement for \textsf{ps:propName}.

% Qualified Scoped Functionality
\noindent\textbf{Qualified Scoped Functionality:} ``A Predicate Statement about a Subject refers to at most one Object.''
\begin{align}
  \textsf{ex:Sub} &\sqsubseteq \mathord{\leq}1 \textsf{p:propName.}\top \\
  \textsf{ex:Sub} &\sqsubseteq \mathord{\leq}1 \textsf{wdt:propName.ex:Sub} \\
  \textsf{ex:Sub} &\sqsubseteq \mathord{\leq}1 \textsf{ps:propName.ex:Obj}
\end{align}
We need to include the third axiom which scopes the global functionality statement for \textsf{ps:propName}.

% Inverse Qualified Scoped Functionality
\noindent\textbf{Inverse Qualified Scoped Functionality:} ``A Predicate Statement that refers to an Object is about at most one Subject.''
\begin{align}
  \textsf{ex:Obj} &\sqsubseteq \mathord{\leq}1 \textsf{ps:propName-.}\top \\
  \textsf{ex:Obj} &\sqsubseteq \mathord{\leq}1 \textsf{wdt:propName-.ex:Sub} \\
  \textsf{wb:Statement}   &\sqsubseteq \mathord{\leq}1 \textsf{p:propName-.ex:Sub}
\end{align}

% Existential
\noindent\textbf{Existential:} ``A Predicate Statement refers to at least one Object.''
\begin{align}
  \textsf{ex:Sub} &\sqsubseteq \exists \textsf{p:propName.}\top \\
  \textsf{ex:Sub} &\sqsubseteq \exists \textsf{wdt:propName.ex:Obj}
\end{align}

% Inverse Existential
\noindent\textbf{Inverse Existential:} ``A Predicate Statement is about at least one Subject.''
\begin{align}
  \textsf{ex:Obj} &\sqsubseteq \exists \textsf{ps:propName.}\top \\
  \textsf{ex:Obj} &\sqsubseteq \exists \textsf{wdt:propName}^-\textsf{.ex:Sub}
\end{align}
Note that Inverse Existential axioms will only work when in conjunction with a domain restriction axiom. In essence, \textsf{p:propName} is inverse functional, but because we do not have any control over \textsf{wb:Statement}, we cannot state that \textsf{p:propName} is also inverse existential, as that would interfere every other \textsf{wb:Statement}. As such, we can state that \textsf{ps:propName} is inverse existential, which means that there exists a \textsf{wb:Statement} node, and we can then assume the existence of something that points at that node. Yet we cannot simultaneously dictate the type of that node. However, if and only if we have domain restriction axiom for \textsf{p:propName}, we can together with the other axioms approximate Inverse Existentiality.

%%%%%%%%%%%%%%%%%%%%%%%%%%%%%%%%%%%%%%%%%%%%%%%%%%%%%%%%%%%%%%%%
\section{Case Study: The Enslaved Ontology}
\label{sec:example}
%%%%%
\begin{figure*}[tbp]
    \centering
    \includegraphics[width=\textwidth]{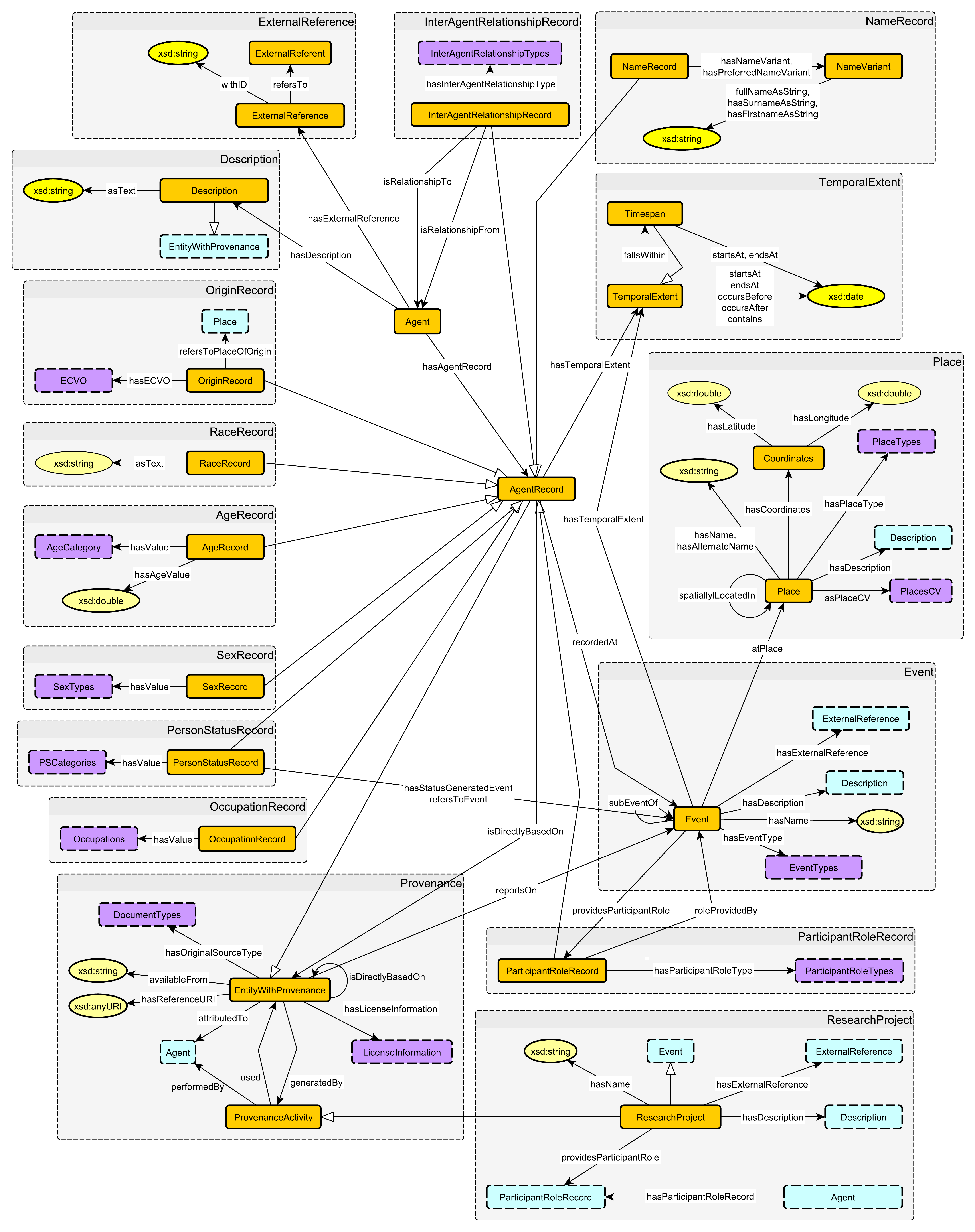}
    \caption{The original Enslaved Ontology}
    \label{fig:eo-orig}
\end{figure*}
\begin{figure*}[tbp]
    \centering
    \includegraphics[width=\textwidth, angle=90]{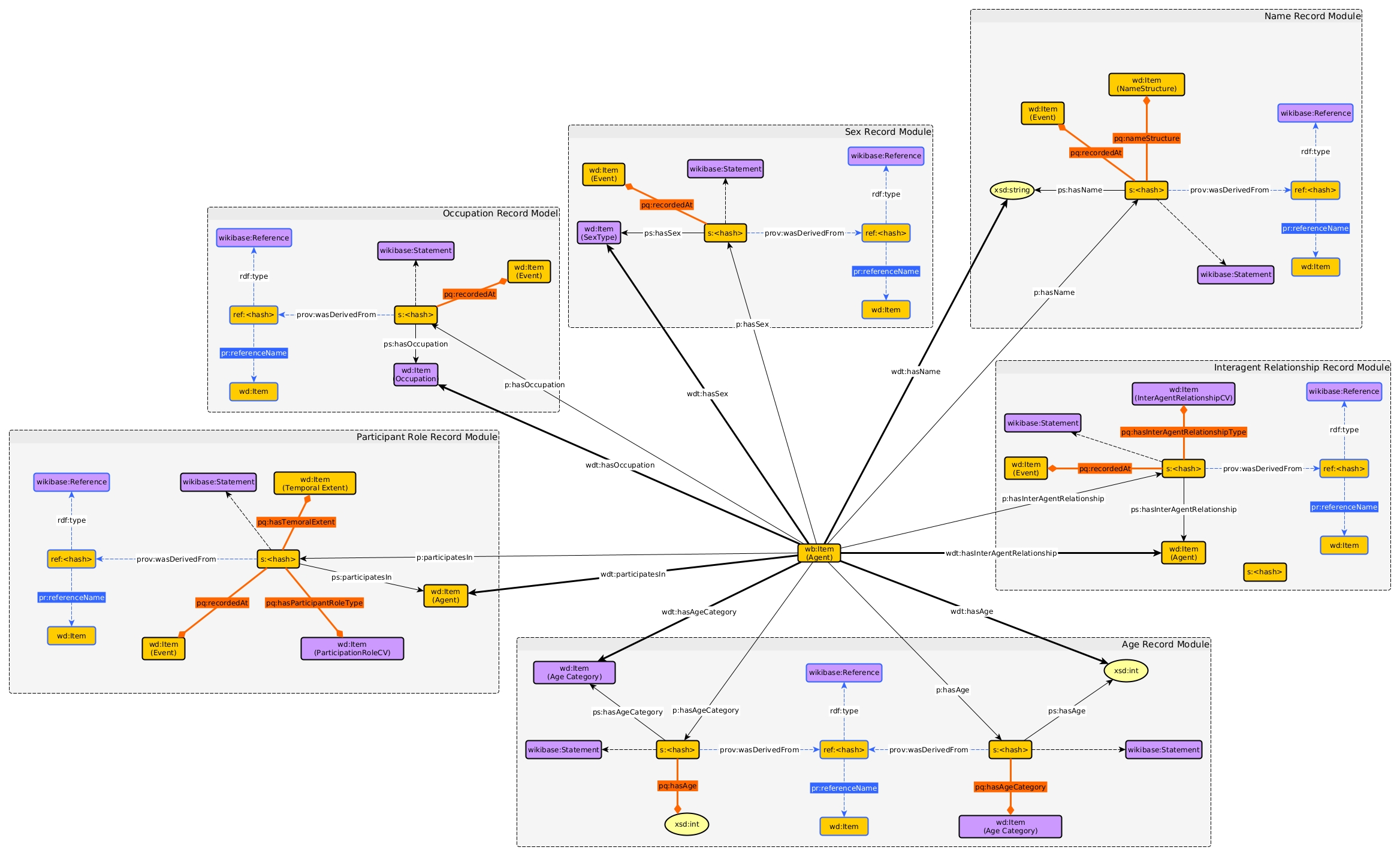}
    \caption{A reconstruction of some of the modules from the Enslaved Ontology as mapped into the Wikibase model. Recall that purple rounded rectangles indicate that that class is controlled, i.e., as in a controlled vocabulary.}
    \label{fig:eo-recon-expanded}
\end{figure*}
\begin{figure*}[tbp]
    \centering
    \includegraphics[width=\textwidth]{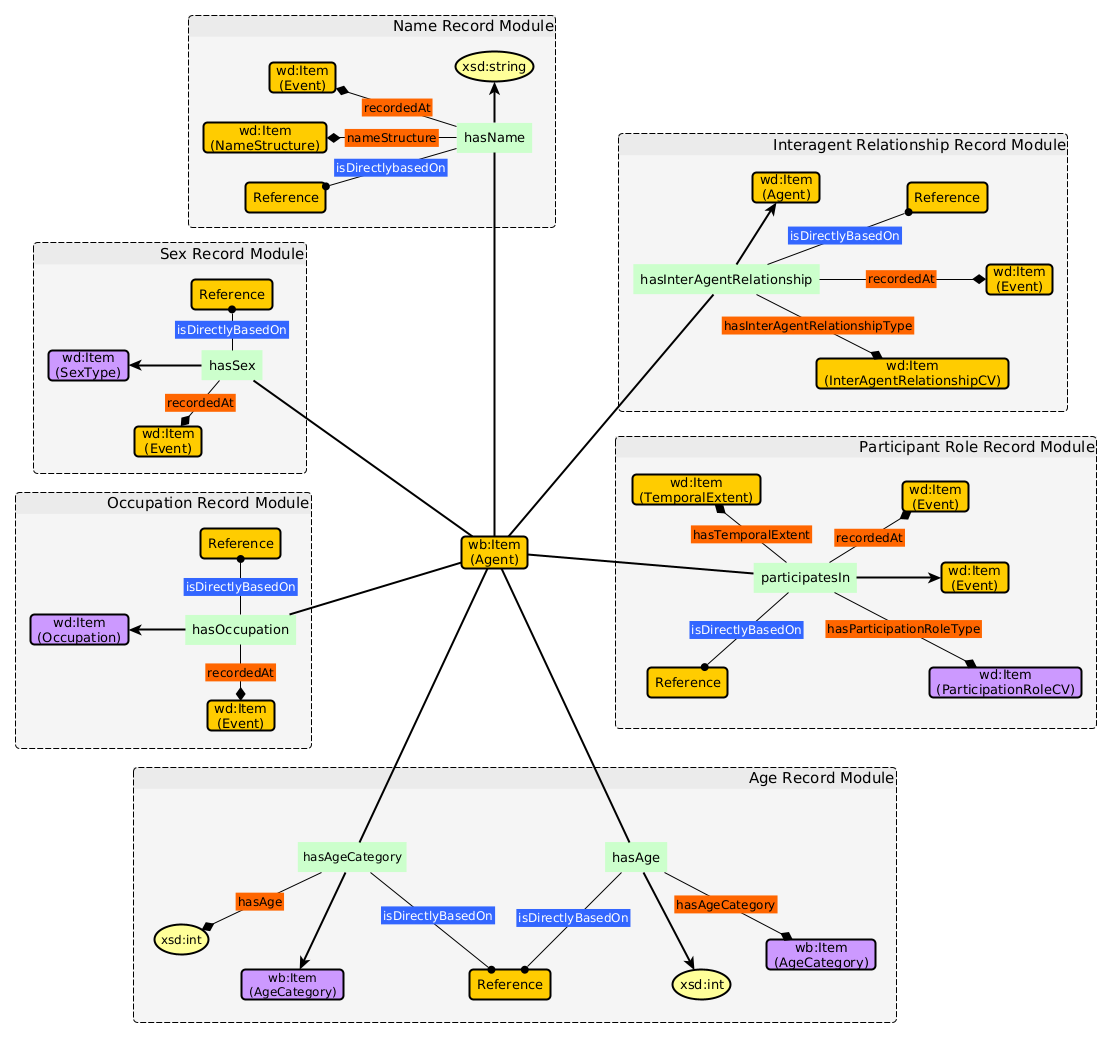}
    \caption{A reconstruction of some of the modules from the Enslaved Ontology using the Wikibase patterns and modified graphical syntax. Recall that purple rounded rectangles indicate that that class is controlled, i.e., as in a controlled vocabulary.}
    \label{fig:eo-recon}
\end{figure*}
%%%%%%%%%%%%%%%%%%%%%%%%%%%%%%%%%%%%%%%%%%%%%%%%%%%%%%%%%%%%%%%%
\subsection{The Enslaved Ontology}
%%%%%
The Enslaved Ontology serves as the underlying schema and data organization paradigm for the Enslaved Hub. It is not used for reasoning or inference, but as a guide for organizing and integrating the data, and understanding the knowledge base as a whole. As previously discussed, the Enslaved Ontology was developed using a nascent version of the MOMo Methodology and, furthermore, before the decision to use Wikibase as the underlying implementation and infrastructure for serving the data.

The work described herein is a result of the mismatches between the original Enslaved Ontology (whose schema diagram is shown in Figure \ref{fig:eo-orig}) and how Wikibase stores information.

We have reconstructed the Enslaved Ontology using our modified graphical syntax, resulting in Figure \ref{fig:eo-recon}. The individual modules appear in Figures \ref{fig:sex-rec} through \ref{fig:age-rec}. At this time, we have only reconstructed some of the modules, with the remaining modules relegated to future work.

%%%%%%%%%%%%%%%%%%%%%%%%%%%%%%%%%%%%%%%%%%%%%%%%%%%%%%%%%%%%%%%%
\section{Related Work}
\label{sec:related}
%%%%%

\subsection*{CIDOC-CRM}
The CIDOC conceptual reference model (CIDOC-CRM) is an informational model for representing cultural information \cite{cidoc}. As mentioned in Section \ref{sec:intro}, we are particularly interested in persistence of data, but also facilitating robust deployment of interacting with the data. While CIDOC-CRM is a domain-standard way of annotating data, which improves the interoperability of the data with other similarly described cultural data, we would yet need to design a system capable of serving that data. It is for this reason we initially chose to align to the Wikibase model. Creating conceptual patterns between Wikibase and CIDOC-CRM is potential future work.

\subsection*{OTTR}
Reasonable Ontology Design Templates (OTTR) \cite{ottr} is a methodology for designing \emph{templates} for concepts. A tutorial can be found online.\footnote{https://ottr.xyz/} In particular, it allows for the schema developer to design a base level template for a certain concept, which can then be easily and programmatically expanded into the appropriate axiomatization. However, it cannot currently instantiate from the property graph formulation (our conceptual diagrams) to the expansions. Extending OTTR to work this way, or identifying a sufficiently capable workaround, is also potential future work.

\subsection*{Property Graphs}
Property graphs \cite{prop-graphs} allow for the specification of predicates as ``first-class citizens.'' While this is a natural way of attaching qualifiers and references to the \textsf{wdt:propertyName}s, it is not currently possible to specify the more interesting axioms (such as domain and range restrictions) in OWL over such graphs. We look forward to evaluating how RDF* and SPARQL* will be formalized from the upcoming W3C working group, which may provide an additional way of modeling such data. It still remains to be seen how semantics may be expressed over such structures.

\subsection*{Open Data to Wikidata}
% Patterns
In \cite{od2wd}, the authors take a pattern-based approach to semi-automatically populating Wikidata from open (tabular) data. This is similar in purpose to our work: persist data in a transparent manner and utilize the Wikibase model. However, it significantly departs from our work; foremost is that it is an automatic framework that creates a naive schema from tabular data and attempts to match these entities, and subsequent instance data, to existing entities in Wikidata. This is a departure from our approach which is to create a conceptual pattern library for the development of rich schemas.

%%%%%%%%%%%%%%%%%%%%%%%%%%%%%%%%%%%%%%%%%%%%%%%%%%%%%%%%%%%%%%%%
\section{Conclusion}
\label{sec:conc}
%%%%%
% Taken from wop paper
When developing and deploying a knowledge graph, there are many obstacles to a persistent, transparent, and usable resource. One way to overcome these obstacles is to use the Wikibase framework. In this paper, we have represented several common modeling constructions in a graphical syntax that makes it clear how they map into the Wikibase context. This should allow ontology developers to more quickly, accurately, and with reduced effort create ontologies (or knowledge graph schema) that are ``Wikibase ready,'' thus improving persistence and accessibility of the deployed knowledge graph. 

%%%%%%%%%%%%%%%%%%%%%%%%%%%%%%%%%%%%%%%%%%%%%%%%%%%%%%%%%%%%%%%%
\subsection*{Future Work}
%%%%%
There is certainly additional work to be accomplished in this direction. In particular, we see the following as immediate next steps to take.
\begin{compactitem}
    \item Identification of frequent CIDOC \cite{cidoc} patterns and corresponding translation into Wikibase patterns.
    \item Extension of OTTR \cite{ottr} to allow for instantiations from the conceptual diagrams.
    \item Develop a robust or extend a tooling system (e.g., CoModIDE \cite{momo-swj} for directly using these ``Wikibase-ified'' axiom patterns.
    \item Create a MODL \cite{modl} of Wikibase-compatible patterns (e.g., by taking each pattern in MODL 1.0 and translating them using the axiom patterns above).
\end{compactitem}

%%%%%%% Acks
\medskip
\noindent\emph{Acknowledgement.} The authors acknowledge support by the National Science Foundation under Grant 2032628 \emph{EAGER: Open Science in Semantic Web Research} and Grant 2033521 A1: KnowWhereGraph: Enriching and Linking Cross-Domain Knowledge Graphs using Spatially-Explicit AI Technologies, as well as the Mellon Foundation through the \emph{Enslaved: Peoples of the Historical Slave Trade}. 
%%%%%%%%%%%%%%%%%%%%%%%%%%%%%%%%%%%%%%%%%%%%%%%%%%%%%%%%%%%%%%%%
\bibliographystyle{abbrv}
\bibliography{refs}
%%%%%%%%%%%%%%%%%%%%%%%%%%%%%%%%%%%%%%%%%%%%%%%%%%%%%%%%%%%%%%%%
%%%%%%%%%%%%%%%%%%%%%%%%%%%%%%%%%%%%%%%%%%%%%%%%%%%%%%%%%%%%%%%%
\end{document}